\documentclass{article} 
\usepackage{iclr2022_conference,times}

\usepackage{amsmath,amsfonts,bm}

\def\eqref#1{equation~\ref{#1}}

\def\1{\bm{1}}

\DeclareMathAlphabet{\mathsfit}{\encodingdefault}{\sfdefault}{m}{sl}
\SetMathAlphabet{\mathsfit}{bold}{\encodingdefault}{\sfdefault}{bx}{n}

\usepackage{hyperref}
\usepackage{url}

\usepackage{graphicx}
\usepackage{subcaption}
\usepackage{booktabs}
\usepackage{wrapfig, blindtext, lipsum}
\usepackage{makecell}
\usepackage{comment}
\usepackage{colortbl}

\title{MobileViTv3: Mobile-Friendly Vision Transformer with Simple and Effective Fusion of Local, Global and Input Features}

\author{Shakti N. Wadekar\thanks{This work was completed as an Intern at Micron Technology Inc.} \\
\texttt{swadekar@purdue.edu} \\
\And
Abhishek Chaurasia \\
\texttt{achaurasia@micron.com} \\
}

\iclrfinalcopy 
\begin{document}

\maketitle

\begin{abstract}
MobileViT (MobileViTv1) combines convolutional neural networks (CNNs) and vision transformers (ViTs) to create light-weight models for mobile vision tasks.
Though the main MobileViTv1-block helps to achieve competitive state-of-the-art results, the fusion block inside MobileViTv1-block, creates scaling challenges and has a complex learning task.
We propose changes to the fusion block that are \textit{simple and effective} to create MobileViTv3-block, which addresses the scaling and simplifies the learning task. 
Our proposed MobileViTv3-block used to create MobileViTv3-XXS, XS and S models outperform MobileViTv1 on ImageNet-1k, ADE20K, COCO and PascalVOC2012 datasets. 
On ImageNet-1K, MobileViTv3-XXS and MobileViTv3-XS surpasses MobileViTv1-XXS and MobileViTv1-XS by 2\% and 1.9\% respectively.
Recently published MobileViTv2 architecture removes fusion block and uses linear complexity transformers to perform better than MobileViTv1. 
We add our proposed fusion block to MobileViTv2 to create MobileViTv3-0.5,0.75 and 1.0 models.
These new models give better accuracy numbers on ImageNet-1k, ADE20K, COCO and PascalVOC2012 datasets as compared to MobileViTv2.
MobileViTv3-0.5 and MobileViTv3-0.75 outperforms MobileViTv2-0.5 and MobileViTv2-0.75 by 2.1\% and 1.0\% respectively on ImageNet-1K dataset.
For segmentation task, MobileViTv3-1.0 achieves 2.07\% and 1.1\% better mIOU compared to MobileViTv2-1.0 on ADE20K dataset and PascalVOC2012 dataset respectively. 
Our code and the trained models are available at \href{https://github.com/micronDLA/MobileViTv3}{https://github.com/micronDLA/MobileViTv3}.
\end{abstract}

\section{Introduction}

Convolutional Neural Networks(CNNs) [like ResNet \citep{he2016deep}, DenseNet \citep{huang2017densely} and EfficientNet \citep{tan2019efficientnet}] are widely used for vision tasks such classification, detection and segmentation, due to their strong performance on the established benchmark datasets such as Imagenet \citep{russakovsky2015imagenet}, COCO \citep{lin2014microsoft}, PascalVOC \citep{Everingham15}, ADE20K \citep{zhou2017scene} and other similar datasets. 
When deploying CNNs on edge devices like mobile phones which are generally resource constrained, light-weight CNNs suitable for such environments come from family of models of MobileNets  (MobileNetv1, MobileNetv2, MobileNetv3) \citep{howard2019searching}, ShuffleNets (ShuffleNetv1 and ShuffleNetv2) \citep{ma2018shufflenet} and light weight versions of EfficientNet \citep{tan2019efficientnet} (EfficientNet-B0 and EfficientNet-B1). 
These light-weight models lack in accuracy when compared to models with large parameters and FLOPs. 
Recently, Vision Transformers(ViTs) have emerged as an strong alternative to CNNs on these vision tasks. 
CNNs, due to its architecture design, interacts with local neighbouring pixels/feature to produce feature maps which has local information embedded in them.
In contrast, self-attention mechanism in ViTs interacts with all parts of the image/feature map to produce features which have global information embedded in them. 
This has been demonstrated to produce comparable results to CNNs but with large pre-training data and advance data augmentation \citep{dosovitskiy2020image}.   
Also, this global processing comes at a cost of large parameters and FLOPs to match the performance of CNNs as seen in ViT \citep{dosovitskiy2020image}, and its different versions such as DeiT \citep{touvron2021training}, SwinT \citep{liu2021swin}, MViT \citep{fan2021multiscale}, Focal-ViT \citep{yang2021focal}, PVT \citep{wang2021pyramid}, T2T-ViT \citep{yuan2021tokens}, XCiT \citep{ali2021xcit}. 
\citep{xiao2021early} demonstrates that ViTs suffer from problem of high sensitivity to hyperparameters such as choice of optimizer, learning rate, weight decay and slow convergence.
To address these problems \citep{xiao2021early} proposes to introduce convolutional layers in ViTs.

\begin{figure}[t]
    \begin{center}
    \includegraphics[width=13.5cm]{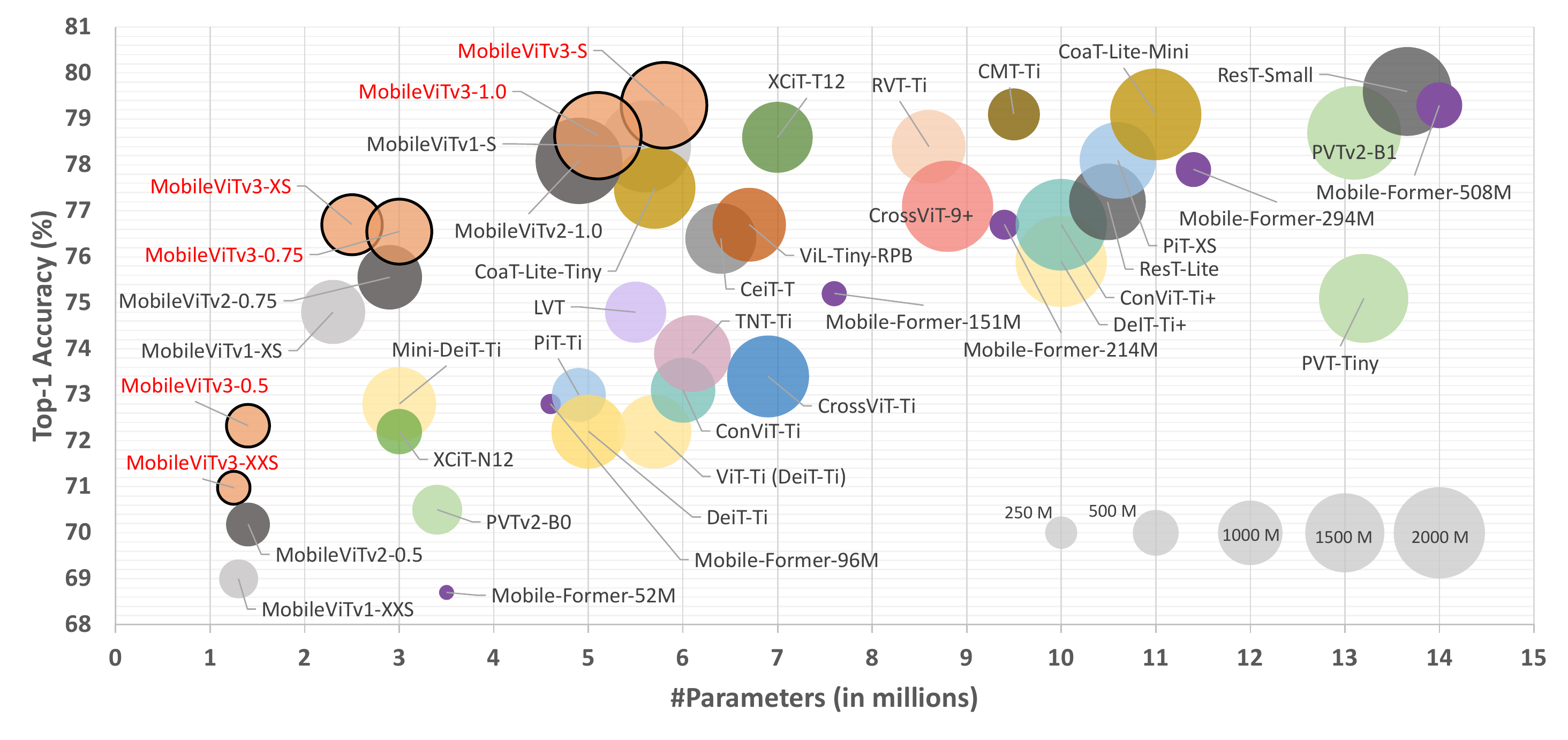}
    \end{center}
    \caption{Comparing Top-1 accuracies of MobileViTv3, ViT variants and hybrid models on ImageNet-1K dataset. The area of bubbles correspond to number of FLOPs in the model. The reference FLOP sizes are shown in the bottom right (example, 250M is 250 Mega-FLOPs/Million-FLOPs). Models of our MobileViTv3 architecture outperforms other models with similar parameter budget of under 2M, 2-4M and 4-8M. Also, they achieve competitive results when compared to the models greater than 8M parameters.}
    \label{fig:ViT_comp}
\end{figure}

Many recent works have introduced convolutional layers in ViT architecture to form hybrid networks to improve performance, achieve sample efficiency and make the models more efficient in terms of parameters and FLOPs like MobileViTs (MobileViTv1 \citep{mehta2021mobilevit}, MobileViTv2\citep{mehta2022mobilevit}), CMT \citep{guo2022cmt}, CvT \citep{wu2021cvt}, PVTv2 \citep{wang2022pvt}, ResT \citep{zhang2021rest}, MobileFormer \citep{chen2022mobile}, CPVT \citep{chu2021conditional}, MiniViT \citep{zhang2022minivit}, CoAtNet \citep{dai2021coatnet}, CoaT \citep{xu2021co}. 
Performance of many of these models on ImageNet-1K, with parameters and FLOPs is shown in Figure  \ref{fig:ViT_comp}.
Among these models, only MobileViTs and MobileFormer are specifically designed for resource constrained environment such as mobile devices.
These two models achieve competitive performance compared to other hybrid networks with less parameters and FLOPs.
Even though these small hybrid models are critical for the vision tasks on mobile devices, there is little work done in this area.

Our work focuses on improving one such light-weight family of models known as MobileViTs (MobileViTv1 \citep{mehta2021mobilevit} and MobileViTv2 \citep{mehta2022mobilevit}).
When compared to the models with parameter budget of 6 million(M) or less, MobileViTs achieve competitive state-of-the-art results with a simple training recipe (basic data augmentation) on classification task.
Also it can be used as an efficient backbone across different vision tasks such as detection and segmentation.
While focusing on only the models with 6M parameters or less, we pose the question:
Is it possible to change the model architecture to improve its performance by maintaining similar parameters and FLOPs?
\textit{To do so, our work looks into challenges of MobileViT-block architecture and proposes \textbf{simple and effective} way to fuse input, local (CNN) and global (ViT) features which lead to significant performance improvements on Imagenet-1K, ADE20k, PascalVOC and COCO dataset}. 
We propose four main changes to MobileViTv1 block (three changes w.r.t MobileViTv2 block) as shown in figure \ref{fig:blocks}.
Three changes are in the fusion block:
First, 3x3 convolutional layer is replaced with 1x1 convolutional layer.
Second, features of local and global representation blocks are fused together instead of input and global representation blocks. 
Third, input features are added in the fusion block as a final step before generating the output of MobileViT block.
Fourth change is proposed in local representation block where normal 3x3 convolutional layer is replaced by depthwise 3x3 convolutional layer. 
These changes result in the reduction of parameters and FLOPs of MobileViTv1 block and allow scaling (increasing width of the model) to create a new MobileViTv3-S, XS and XXS architecture, which outperforms MobileViTv1 on classification (Figure \ref{fig:ViT_comp}), segmentation and detection tasks.
For example, MobileViTv3-XXS and MobileViTv3-XS perform 2\% and 1.9\% better with similar parameters and FLOPs on ImageNet-1K dataset compared to MobileViTv1-XXS and MobileViTv1-XS respectively.
In MobileViTv2, fusion block is absent. Our proposed fusion block is introduced in MobileViTv2 architecture to create  MobileViTv3-1.0, 0.75 and 0.5 architectures. 
MobileViTv3-0.5 and MobileViTv3-0.75 outperforms MobileViTv2-0.5 and MobileViTv2-0.75 by 2.1\% and 1.0\% respectively with similar parameters and FLOPs on ImageNet-1K dataset.

\section{Related Work}

\textbf{Vision Transformers}: 
ViT \citep{dosovitskiy2020image} introduced the transformer models used for Natural Language Processing tasks to vision domain, specifically for image recognition. 
Later, its different versions such as DeiT \citep{touvron2021training} improved the performance by introducing a novel training technique and reducing the dependency on large pre-training data.  
Works focusing on improving self-attention mechanism to boost performance include XCiT \citep{ali2021xcit}, SwinT \citep{liu2021swin}, ViL \citep{zhang2021multi} and Focal-transformer \citep{yang2021focal}.
XCiT introduces cross-covariance attention where self-attention is operated on feature channels instead of tokens and interactions are based on cross-covariance matrix between keys and queries.
SwinT modified the ViT to make it a general purpose architecture which can be used for various vision tasks like classification, detection and segmentation. 
This was achieved by replacing self-attention mechanism with a shifted-window based self-attention which allows model to be adapted for different input image scales and do it efficiently by achieving a linear computational complexity relation with the input image size.
ViL improves ViT by encoding image at multiple scales and uses self-attention mechanism which is a variant of Longformer \citep{beltagy2020longformer}.
Recent works like T2T-ViT \citep{yuan2021tokens} and PVT (PVTv1) \citep{wang2021pyramid}  also focus on introducing CNN like hierarchical feature learning by reducing spatial resolution or token sizes of output after each layer.
T2T-ViT proposes layer-wise token-to-token transformation where neighbouring tokens are recursively aggregated into one token to capture local structure and reduce token lengths.
PVT a pyramid vision transformer, successively reduces the feature map resolution size reducing computational complexity and achieving competitive results on ImageNet-1K.
Few new architectures like CrossViT \citep{chen2021crossvit}, MViT \citep{fan2021multiscale}, MViTv2  \citep{li2022mvitv2} and Focal-transformer  \citep{yang2021focal} learn both local features (features learnt specifically from neighbouring pixels/features/patches) and global features (features learnt using all pixels/features/patches).
Focal-transformer replaces self-attention with focal-self-attention where each token is able to attend its closest surrounding tokens at fine granularity and also tokens far away at coarse granularity, capturing both short and long-range visual dependencies.
CrossViT processes small-patch and large-patch tokens separately and fused together by attention multiple times to complement each other.
MViT designed for video and image recognition, learns multi-scale pyramid of features where early layers capture low-level visual information and deeper layers capture complex and high dimensional features. 
MViTv2 further improves MViT by incorporating positional embedding and residual pooling connections in its architecture.

\textbf{CNNs}: 
ResNet \citep{he2016deep} is one of the most widely used general purpose architecture for vision tasks like classification, segmentation and detection. 
ResNet architecture, due to its residual connections, helps optimization of deeper layers, allowing construction of deep neural networks (deep CNNs). These deep CNNs are able to achieve state of the art results on various benchmarks.
DenseNet \citep{huang2017densely} inspired from ResNet, connects every layer to the next using skip connection in a feed-forward fashion. 
Other CNNs like ConvNeXt \citep{liu2022convnet}, RegNetY \citep{radosavovic2020designing}, SqueezeNet \citep{iandola2016squeezenet} and Inception-v3 \citep{szegedy2016rethinking} also have achieved competitive state-of-the-art performance. 
But, the best performing CNN models are generally high in number of parameters and FLOPs.
Light-weight CNNs that achieve competitive performance with less parameters and FLOPs include EfficientNet  \citep{tan2019efficientnet}, MobileNetV3  \citep{howard2019searching}, ShuffleNetv2  \citep{ma2018shufflenet} and ESPNetv2  \citep{mehta2019espnetv2}.
EfficientNet studied model scaling and developed family of efficientnet models which are still one of the most efficient CNNs in terms of parameters and FLOPs. 
MobileNetV3 belongs to category of models specifically developed for resource constrained environments such as Mobile phones. 
Building block of MobileNetV3 architecture uses MobileNetv2 \citep{sandler2018mobilenetv2} block and Squeeze-and-Excite \citep{hu2018squeeze} network in it. 
ShuffleNetv2 studies and proposes guidelines for efficient model design and produces shufflenetv2 family of models which also performs competitively with other light-weight CNN models.
ESPNetv2 uses depth-wise dilated separable convolution to create EESP (Extremely Efficient Spatial Pyramid) unit which helps to reduce parameters and FLOPs and achieve competitive results.

\textbf{Hybrids}: 
Recently, many different models are being proposed which combines CNNs and ViTs together in one architecture to capture both long range dependencies using self-attention mechanism of ViT and  local information using local kernels in CNNs to improve performance on vision tasks. 
MobileViT (MobileViTv1, MobileViTv2) \citep{mehta2021mobilevit} and MobileFormer \citep{chen2022mobile} have been specifically designed for constrained environments like mobile devices. 
MobileViTv1 and MobileViTv2 achieve state-of-the-art results when compared to models with a parameter budget of 6M or less. 
MobileFormer architecture combines MobileNetv3 and ViT to also achieve competitive results. 
CMT \citep{guo2022cmt} architecture has convolutional stem, convolutional layer before every transformer block and stacks convolutional layers and transformer layers alternatively. 
CvT \citep{wu2021cvt} uses convolutional token embedding instead of linear embedding used in ViTs and a convolutional transformer layer block that leverages these convolutional token embeddings to improve performance. 
PVTv2 \citep{wang2022pvt} in contrast to PVTv1 uses convolutional feed forward layers in transformers, overlapping patch embedding and linear complexity attention layer to gain improvements over PVT. 
ResT \citep{zhang2021rest} uses depth-wise convolution in self-attention (for memory efficiency) and patch embedding as stack of overlapping convolution operation with stride on token map. 
CoAtNet \citep{dai2021coatnet} unifies depthwise convolution and self-attention using simple relative attention, also vertically stacks convolutional layers and attention layers. 
PiT's \citep{heo2021rethinking} pooling layer uses depthwise convolution to achieve spatial reduction for boosting performance. 
LVT \citep{yang2022lite} introduces convolutional self-attention where local self-attention is introduces within a convolutional kernel and also recursive atrous self-attention to encompass multi-scale context  to improve performance. 
ViTAE \citep{xu2021vitae} has convolution layers in parallel to multi-head self-attention module and both are fused and fed to feedforward network, also ViTAE uses convolutional layers to embed inputs to token.
CeiT \citep{yuan2021incorporating} introduces locally enhanced feed-forward by using depth-wise convolution with other changes to achieve competitive results. 
RVT \citep{mao2022towards} uses convolutional stem to generate patch embeddings and uses convolutional feed-forward network in transformer to achieve better results.  

\section{New MobileViT architecture}
\label{headings}

\begin{figure}
\centering
\begin{subfigure}{.5\textwidth}
  \centering
  \includegraphics[width=1.0\linewidth]{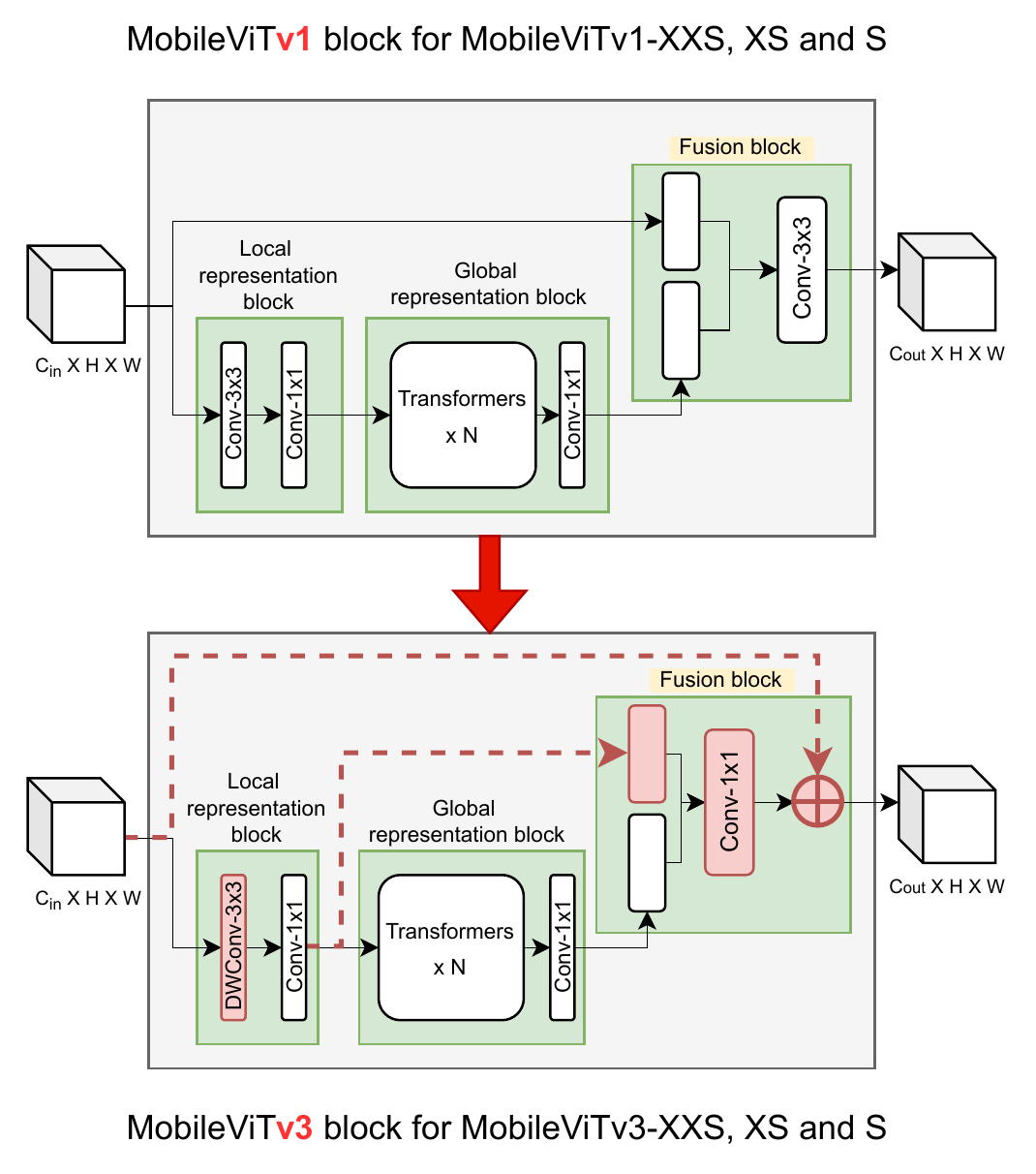}
  \captionof{figure}{MobileViTv1 and MobileViTv3 blocks}
  \label{fig:mv1v3}
\end{subfigure}%
\begin{subfigure}{.5\textwidth}
  \centering
  \includegraphics[width=1.0\linewidth]{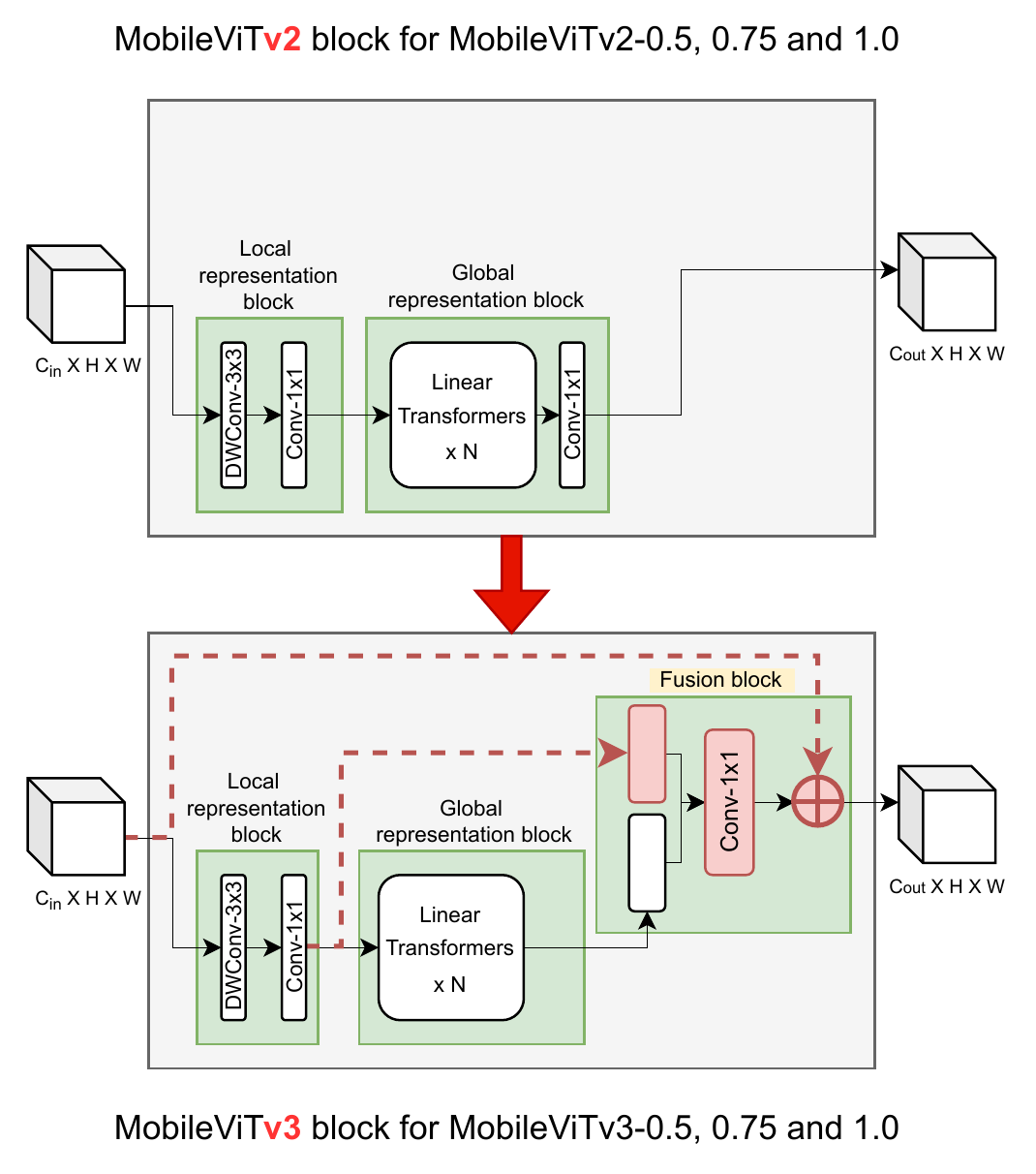}
  \captionof{figure}{MobileViTv2 and MobileViTv3 blocks}
  \label{fig:mv2v3}
\end{subfigure}
\caption{A comparison between: (a) MobileViTv1 and MobileViTv3 modules, and (b) MobileViTv2 and MobileViTv3 modules. The proposed architectural changes are highlighted in red.}
\label{fig:blocks}
\end{figure}

Our work proposes four design changes to the existing MobileViTv1 block architecture to build MobileViTv3-block as shown in Figure \ref{fig:mv1v3}. 
Section \ref{section:mv3blocks} explains these four changes in MobileViTv3-block architecture and compares with MobileViTv1 and MobileViTv2-blocks. 
Section \ref{section:scaleup} details MobileViTv3-S, XS and XXS architectures and shows how it is scaled compared to MobileViTv1-S, XS and XXS.
In recently published MobileViTv2 architecture, changes applied to MobileViT-block are, fusion block is removed, transformer block uses self-attention with linear complexity and depthwise convolutional layer is used in local representation block. We add back the fusion block with our proposed changes to create MobileViTv3-block as shown in Figure \ref{fig:mv2v3} for MobileViTv3-0.5, 0.75, 1.0 architectures.

\subsection{MobileViTv3 block} \label{section:mv3blocks}



\textbf{Replacing 3x3 convolutional layer with 1x1 convolutional layer in fusion block}: 
Two main motivations exist for replacing 3x3 convolutional layer in fusion.
\textit{First, fuse local and global features independent of other locations in the feature map to simplify the fusion block's learning task}. 
Conceptually, 3x3 convolutional layer is fusing input features, global features, and other location's input and global features which are present in the receptive field, which is a complex task. 
Fusion block's goal can be simplified by allowing it to fuse input and global features, independent of other locations in feature map. 
To do so, we use 1x1 convolutional layer in fusion instead of 3x3 convolutional layer.
\textit{Second, is to remove one of the major constraints in scaling of MobileViTv1 architecture}. 
Scaling MobileViTv1 from XXS to S is done by changing width of the network and keeping depth constant. 
Changing width (number of input and output channels) of MobileViTv1 block causes large increase in number of parameters and FLOPs. 
For example, if the input and output channels are doubled (2x) in MobileViTv1 block, the number of input channels to 3x3 convolutional layer inside fusion block increases by 4x and output channels by 2x, because input to the 3x3 convolutional layer is concatenation of input and global representation block features. This causes a large increase in parameters and FLOPs of MobileViTv1 block. Using 1x1 convolutional layer avoids this large increase in parameters and FLOPs while scaling.
 

\textbf{Local and Global features fusion}: 
In fusion layer, features from \textit{local} and global representation blocks are concatenated in our proposed MobileViTv3 block instead of \textit{input} and global representation features.
This is because the local representation features are more closely related to the global representation features when compared to the input features.
The output channels of the local representation block are slightly higher than the channels in input features.
This causes an increase in the number of input feature maps to the fusion block's 1x1 convolutional layer, but the total number of parameters and FLOPs are significantly less than the baseline MobileViTv1 block due to the change of 3x3 convolutional layer to 1x1 convolutional layer.

\textbf{Fusing input features}: 
Input features are added to the output of 1x1 convolutional layer in the fusion block. 
The residual connections in models like ResNet and DenseNet have shown to help the optimization of deeper layers in the architecture. 
By adding the input features to the output in the fusion block, we introduce this residual connection in new MobileViTv3 architecture.
Ablation study results shown in table \ref{table:ablation} demonstrates that this residual connection contributes 0.6\% accuracy gain.

\textbf{Depthwise convolutional layer in local representation block}: 
To further reduce parameters, 3x3 convolutional layer in local representation block is replaced with depthwise 3x3 convolutional layer.
As seen in the ablation study results table \ref{table:ablation}, this change does not have a large impact on the Top-1 ImageNet-1K accuracy gain and provides good parameter and accuracy trade-off.

\subsection{Scaling up building blocks} \label{section:scaleup}

Applying the changes proposed in section \ref{section:mv3blocks}, allows scaling of our MobileViTv3 architecture by increasing the width (number of channels) of the layers.
Table \ref{table1:fullarch} shows MobileViTv3-S, XS and XXS architectures with their output channels in each layer, scaling factor, parameters an FLOPs

\begin{table}[h] 
\scriptsize
\centering
\begin{tabular}{lcccccc} \toprule[1.5pt]
    {Layer} & {Size} & {Stride} & {Repeat}& {XXS} & {XS} & {S} \\ 
    \midrule[1.5pt]
    Image          & 256x256  & 1   &   &    &    &    \\
    \midrule
    Conv-3x3, $\downarrow2$  & 128x128   & 2   & 1  & 16    & 16    & 16    \\
    MV2                      & 128x128   & 2   & 1  & 16    & 32    & 32    \\
    \midrule
    MV2     , $\downarrow2$  & 64x64     & 4   & 1  & 24    & 48    & 64    \\
    MV2                      & 64x64     & 4   & 2  & 24    & 48    & 64    \\
    \midrule
    MV2     , $\downarrow2$  & 32x32     & 8   & 1  & 64 (1.3x)  & 96 (1.5x)    & 128 (1.3x)   \\
    MobileViT block (L=2)    & 32x32     & 8   & 1  & 64 (1.3x)  & 96 (1.5x)    & 128 (1.3x)   \\
    \midrule
    MV2     , $\downarrow2$  & 16x16     & 16  & 1  & 80 (1.3x)  & 160 (2.0x)    & 256 (2.0x) \\
    MobileViT block (L=4)    & 16x16     & 16  & 1  & 80 (1.3x)  & 160 (2.0x)    & 256 (2.0x) \\
    \midrule
    MV2     , $\downarrow2$  & 8x8       & 32  & 1  & 128 (1.6x)  & 160 (1.7x)   & 320 (2.0x) \\
    MobileViT block (L=3)    & 8x8       & 32  & 1  & 128 (1.6x)  & 160 (1.7x)   & 320 (2.0x) \\
    \midrule
    Conv-1x1,                & 8x8       & 32  & 1  & 512 (1.6x)   & 640  (1.7x)   & 1280 (2.0x)   \\
    \midrule
    Global pool,             & 1x1       & 256 & 1  & 512   & 640   & 1280   \\
    Linear                   & 1x1       & 256 & 1  & 1000  & 1000  & 1000  \\
    \midrule[1.5pt]
    Parameters (M)           &           &     &    & 1.25   & 2.5   & 5.8   \\
    FLOPs (M)                &           &     &    & 289   & 927   & 1841   \\
    \midrule[1.5pt]
    Top-1 Accuracy (\%)      &           &     &    & 71.0   & 76.7   & 79.3   \\
    
    \bottomrule[1.5pt]
\end{tabular}
\caption{MobileViTv3-S, XS and XXS architecture details and comparison with MobileViTv1-S, XS and XXS. Values given in brackets `()' represent scaling factor compared to MobileViTv1 models. MobileViTv3-XXS, XS and S achieve 2\%, 1.9\% and 0.9\% accuracy gain respectively compared to MobileViTv1-XXS, XS and S on ImageNet-1K dataset.}
\label{table1:fullarch}
\end{table}

\section{Experimental results}
\label{headings}

Our work shows results on classification task using ImageNet-1K in section \ref{results:imgnet}, segmentation task using ADE20K and PASCAL VOC 2012 datasets in section \ref{results:seg}, detection task using COCO dataset in section \ref{results:obj}. 
We also discuss changes to our proposed MobileViTv3 architecture for improving latency and throughput in section \ref{results:latency}.

\subsection{Image Classification on Imagenet-1k} \label{results:imgnet}

\subsubsection{Implementation details} \label{section:classimpdetail}
Except for the batch size, hyperparameters used for MobileViTv3-S, XS and XXS are similar to the MobileViTv1 and hyperparameters used for MobileViTv3-1.0, 0.75 and 0.5 are similar to MobileViTv2. 
Due to resource constraints, we were limited to using a total batch size of 384 (32 images per GPU) for experiments on MobileViTv3-S and XS. To maintain consistency in batch sizes, MobileViTv3-XXS is also trained on batch size of 384.
Batch size of 1020 (85 images per GPU) used of MobileViTv3-0.5,0.75 and 1.0 training.

\textbf{MobileViTv3-S, XS and XXS}: 
Default hyperparameters used from MobileViTv1 include using AdamW as optimizer, multi-scale sampler (S = {(160,160), (192,192), (256,256), (288,288), (320,320)}), learning rate increased from 0.0002 to 0.002 for the first 3K iterations and then annealed to 0.0002 using cosine schedule, L2 weight decay of 0.01, basic data augmentation i.e, random resized cropping and horizontal flipping. 
\textbf{MobileViTv3-1.0, 0.75 and 0.5}: Default hyperparameters used from MobileViTv2 include using AdamW as optimizer, batch-sampler (S = {(256,256)}), learning rate increased from 1e-6 to 0.002 for the first 20K iterations and then annealed to 0.0002 using cosine schedule, L2 weight decay of 0.05, advanced data augmentation i.e, random resized cropping, horizontal flipping, random augmentation, random erase, mixup and cutmix. 
Performance is evaluated using single crop top-1 accuracy, for inference an exponential moving average of model weights is used. 
All the classification models are trained from scratch on the ImageNet-1K classification dataset. 
This dataset contains 1.28M and 50K images for training and validation respectively.

\subsubsection{Comparison with MobileViTs}
Table \ref{table:comp_batch} demonstrates that performance of all the versions of MobileViTv3 surpass MobileViTv1 and MobileViTv2 versions with similar parameters and FLOPs and smaller training batch size. 
The effect of training batch size on MobileViTv3 is also shown in Table \ref{table:comp_batch}.
Increasing total batch size from 192 to 384 improves accuracy of MobileViTv3-S, XS and XXS models. 
This indicates the potential for further accuracy gains with batch size of 1024 on MobileViTv3-XXS, XS and S models. 
It is also important to note that MobileViTv3-S, XS and XXS models trained with basic data augmentation not only outperforms MobileViTv1-S, XS, XXS, but also surpasses performance of MobileViTv2-1.0, 0.75 and 0.5 which are trained with advanced data augmentation. 
Fine-tuned on image size of 384, MobileViTv3-1.0, 0.75 and 0.5 also outperforms fine-tuned versions of MobileViTv2-1.0, 0.75 and 0.5. 

\begin{table}[!thb]
\scriptsize
\centering
\begin{tabular}{cccccc} \toprule
    {Model} & {Training Batch size} & {FLOPs (M)$\downarrow$} & {\# Params. (M)$\downarrow$} & {Top-1 (\%)$\uparrow$} \\ 
    \midrule
    MobileViTv1-XXS  & 1024  & 364 & 1.3 & 69.00 \\
    MobileViTv3-XXS  & 192  & 289  & 1.2 & 70.02 (+1\%)\\
    \textbf{MobileViTv3-XXS}  & 384  & 289  & 1.2 & \textbf{70.98} (+2\%)  \\
    \midrule
    MobileViTv2-0.5  & 1024  & 466 & 1.4 & 70.18 \\
    \textbf{MobileViTv3-0.5}  & 1020  & 481  & 1.4 & \textbf{72.33} (+2.1\%)\\
    \midrule
    MobileViTv2-0.5 (384)  & 64  & 1048 & 1.4 & 72.14 \\
    \textbf{MobileViTv3-0.5} (384)  & 64  & 1083  & 1.4 & \textbf{74.01} (+1.87\%)\\
 \midrule
    MobileViTv1-XS  & 1024  & 986 & 2.3 & 74.8 \\
    MobileViTv3-XS  & 192  & 927  & 2.5 & 76.3 (+1.5\%)  \\
    \textbf{MobileViTv3-XS}  & 384  & 927  & 2.5 & \textbf{76.7} (+1.9\%)  \\
    \midrule
    MobileViTv2-0.75  & 1024  & 1030 & 2.9 & 75.56 \\
    \textbf{MobileViTv3-0.75}  & 1020  & 1064  & 3.0 & \textbf{76.55} (+0.99\%)\\
    \midrule
    MobileViTv2-0.75 (384)  & 64  & 2318 & 2.9 & 76.98 \\
    \textbf{MobileViTv3-0.75} (384)  & 64  & 2395  & 3.0 & \textbf{77.81} (+0.83\%)\\
 \midrule
    MobileViTv1-S  & 1024  & 2009 & 5.6 & 78.4 \\
    MobileViTv3-S  & 192  & 1841  & 5.8 & 78.8 (+0.4\%)  \\ 
    \textbf{MobileViTv3-S}  & 384  & 1841  & 5.8 & \textbf{79.3} (+0.9\%)  \\ 
    \midrule
    MobileViTv2-1.0  & 1024  & 1851 & 4.9 & 78.09 \\
    \textbf{MobileViTv3-1.0}  & 1020  & 1876  & 5.1 & \textbf{78.64} (+0.55\%)\\
    \midrule
    MobileViTv2-1.0 (384)  & 64  & 4083 & 4.9 & 79.68 \\
    \textbf{MobileViTv3-1.0} (384)  & 64  & 4220  & 5.1 & \textbf{79.74} (+0.06\%)\\
    \bottomrule
\end{tabular}
\caption{MobileViT V1, V2 and V3 comparison in terms of Top-1 ImageNet-1k accuracy, parameters and operations. $<$Model name$>$(384) indicate that the model was fine-tuned using an image size of 384x384. Models with similar parameters and operations are grouped together for clear comparison.}
\label{table:comp_batch}
\end{table}

\subsubsection{Comparison with ViTs}
Figure \ref{fig:ViT_comp} compares our proposed MobileViTv3 models performance with other ViT variants and hybrid models. 
Following MobileViTv1, we mainly compare our models with parameter budget of around 6M or less. 
Also, when comparing to models greater than 6M parameters, we limit FLOPs budget to $\sim$2 GFLOPs or less because our largest model in this work has $\sim$2 GFLOPs.

\textbf{Models under 2 million parameters}: 
To the best of our knowledge, only MobileViT variants exist in this range. 
MobileViTv3-XXS and MobileViTv3-0.5 outperform other MobileViT variants. 
MobileViTv3-0.5 by far achieves the best accuracy of 72.33 \%  in 1-2 million parameter budget models (ViT or hybrid). 

\textbf{Models between 2-4 million parameters}: MobileViTv3-XS and MobileViTv3-0.75 outperform all the models in this range. 
Top-1 accuracy of MobileViTv3-XS on ImageNet-1k is 76.7\%, which is 3.9\% higher than Mini-DeiT-Ti \citep{zhang2022minivit}, 4.5 \% higher than XCiT-N12 \citep{ali2021xcit}, and 6.2\% higher than PVTv2-B0 \citep{wang2022pvt}. Although Mobile-Former-53M \citep{chen2022mobile} uses only 53 GFLOPs, it lags in accuracy by a large margin of 12.7\% with MobileViTv3-XS. 

\textbf{Models between 4-8 million parameters}: 
MobileViTv3-S attains the highest accuracy in this parameter range.
MobileViTv3-S with simple training recipe and 300 epochs is 0.7\% better than XCiT-T12 trained using distillation, advanced data augmentation and 400 epochs. 
It is 1.8\%, 2.6\% and 2.9\% better than Coat-Lite-Tiny \citep{xu2021co}, ViL-Tiny-RPB \citep{zhang2021multi} and CeiT-Ti \citep{yuan2021incorporating} respectively. 
MobileViTv3-S is 1\%  better with 0.5x FLOPs and similar parameters as compared to CoaT-Tiny \citep{xu2021co}. 

\textbf{Models greater than 8 million parameters}: We also compare our designed models with existing models having more than 8M parameters and around 2 GFLOPs. 
When compared with MobileViTv3-S trained with basic data augmentation and 300 epochs, CoaT-Lite-Mini achieves competitive accuracy of 79.1\% with $\sim$2x more parameters, similar FLOPs and advanced data augmentation, MobileFormer-508M achieves similar accuracy of 79.3\% with $\sim$2.5x more parameters, $\sim$3.5x less FLOPs, advance data augmentation and 450 training epochs. 
ResT-Small \citep{zhang2021rest} achieves similar accuracy of 79.6\% with $\sim$2.5x more parameters, similar FLOPs and advanced data augmentation. PVTv2-B1 \citep{wang2022pvt} achieves 78.7\% with $\sim$2.3x more parameters, similar FLOPs and advanced data augmentation. CMT-Ti \citep{guo2022cmt}  achieves 79.1\% with $\sim$1.6x more parameters, $\sim$2.9x less FLOPs (due to input image size of 160x160) and advanced data augmentation.

\begin{figure}[t]
\begin{center}
\includegraphics[width=13.5cm]{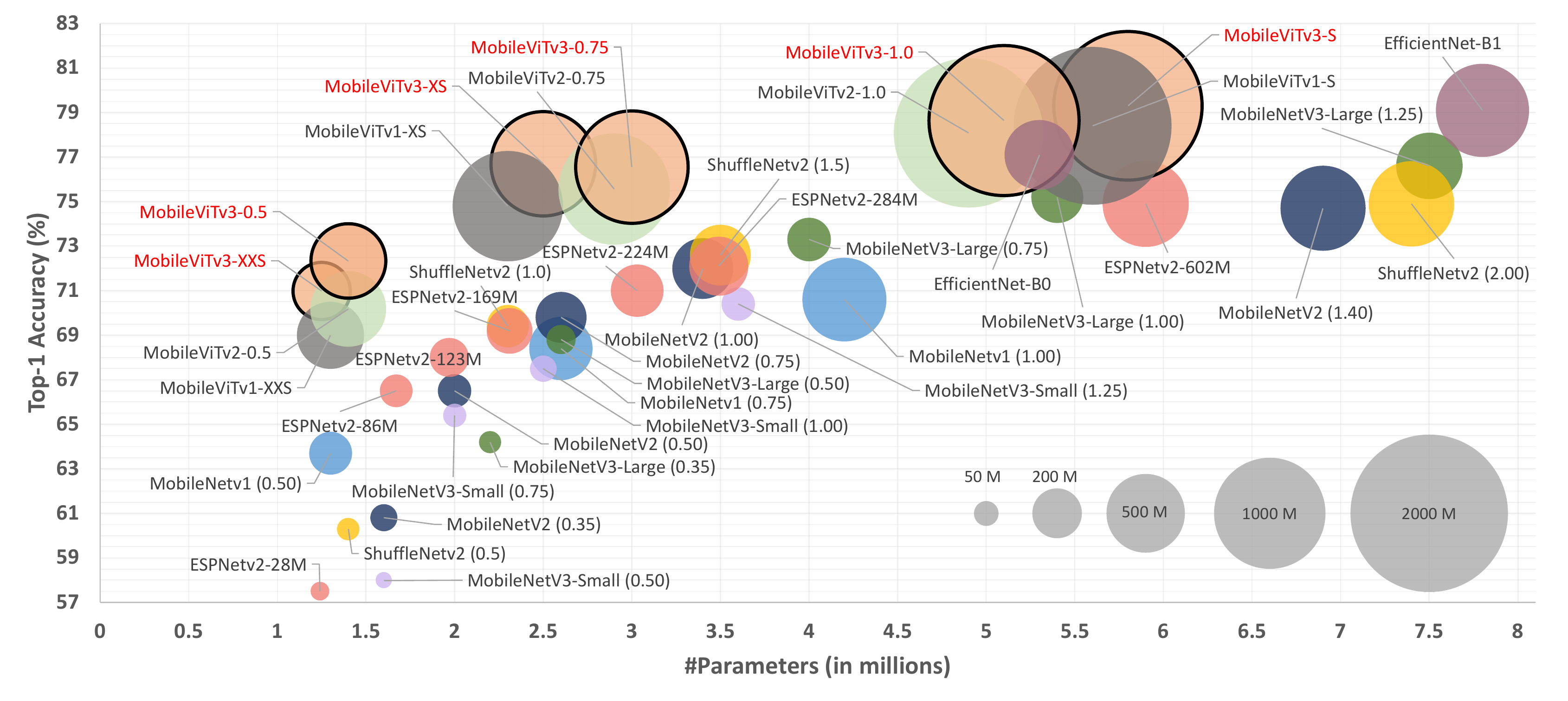}
\end{center}
\caption{Top-1 accuracy comparison between MobileViTv3 models and existing light-weight CNN models on ImageNet-1K dataset. The bubble size corresponds to the number of FLOPs. The reference FLOP sizes are shown in the bottom right (example, 50M is 50 Mega-FLOPs/Million-FLOPs). Models of our MobileViTv3 architecture outperform other models with similar parameter budget of 1-2M, 2-4M and 4-8M.}
\label{fig:comp_cnn}
\end{figure}

\subsubsection{Comparison with CNNs}
Figure \ref{fig:comp_cnn} compares our proposed models with the CNN models which are light-weight with a parameter budget of $\sim$6M or less, similar to MobileViTv1 \citep{mehta2021mobilevit}. 

\textbf{Models in 1-2 million parameters range}:
MobileViTv3-0.5 and MobileViTv3-XXS with 72.33\% and 70.98\% respectively are best accuracies in this parameter range.
MobileViTv3-0.5 achieves over 2.5\% improvement compared to MobileNetv3-small(0.5) \citep{howard2019searching}, MobileNetv3-small(0.75), ShuffleNetv2(0.5) \citep{ma2018shufflenet}, ESPNetv2-28M \citep{mehta2019espnetv2}, ESPNetv2-86M, and ESPNetv2-123M. 

\textbf{Models with 2-4 million parameters}: MobileViTv3-XS achieves over 4\% improvement compared to MobileNetv3-Large(0.75), ShuffleNetv2(1.5), ESPNetv2-284M and MobileNetv2(0.75). 

\textbf{Models with 4-8 million parameters}: MobileViTv3-S shows more than 2\% accuracy gain over EfficientNet-B0 \citep{tan2019efficientnet},  MobileNetv3-Large(1.25), ShuffleNetv2(2.0), ESPNetv2-602M and  MobileNetv2(1.4). 
EfficientNet-B1 with 1.3x more parameters and 2.6x less FLOPs achieves competitive accuracy of 79.1\% compared to MobileViTv3-S with accuracy of 79.3\%.

\subsection{Segmentation} \label{results:seg}

\subsubsection{Implementation details}

\textbf{PASCAL VOC 2012 dataset}: 
Following MobileViTv1, MobileViTv3 is integrated with DeepLabv3 \citep{chen2017rethinking} for segmentation task on PASCAL VOC 2012 dataset \citep{Everingham15}.
Extra annotations and data is used from \citep{hariharan2011semantic} and \citep{lin2014microsoft} respectively, which is a standard practise for training on PascalVOC2012 dataset \citep{chen2017rethinking}; \citep{mehta2019espnetv2}.
For MobileViTv3-S, XS and XXS, training hyperparameters are similar to MobileViTv1 except the batch size. 
Smaller batch size of 48 (12 images per GPU) is used compared to 128 (32 images per GPU) for MobileViTv1. 
Other default hyperparameters include, using adamw as optimizer, weight decay of 0.01, cosine learning rate scheduler, cross-entropy loss and 50 epochs for training. 
For MobileViTv3-1.0, 0.75 and 0.5, all the hyperparameters are kept same as used for MobileViTv2-1.0, 0.75 and 0.5 training. 
Default hyperparameters include using adamw as optimizer, weight decay of 0.05, cosine learning rate scheduler, cross-entropy loss and 50 epochs training. 
The segmentation performance is evaluated on the validation set and reported using mean Intersection over Union (mIOU).

\textbf{ADE20K dataset \citep{zhou2019semantic}}: 
Contains total 25K images with 150 semantic categories. 
Out of 25K images, 20K images are used for training, 3K images for test and 2K images for validation. 
Same training hyperparameters are used for MobileViTv2 models as MobileViTv3-1.0,0.75 and 0.5 models. 
Training hyperparameters include using SGD as optimizer, weight decay of 1e-4, momentum of 0.9, cosine learning rate scheduler, 120 epochs training, cross-entropy loss, batch size 16 (4 images per GPU). 
The segmentation performance is evaluated on the validation set and reported using mean Intersection over Union (mIOU).

\subsubsection{Results}

\textbf{PASCAL VOC 2012 dataset}: 
Table \ref{table:segpascal} demonstrates MobileViTv3 models with lower training batch size of 48, outperforming their corresponding counterpart models of MobileViTv1 and MobileViTv2 which are trained on higher batch size of 128.
MobileViTv3-1.0 achieves 80.04\% mIOU, which outperforms MobileViTv2-1.0 by 1.1\%.
MobileViTv3-XS is  1.6\% better than MobileViTv1-XS and MobileViTv3-0.5 surpasses MobileViTv2-0.5 by 1.41\%. 
\textbf{ADE20K dataset}: Table \ref{table:segade20k} shows the results of MobileViTv3-1.0, 0.75 and 0.5 models on ADE20K dataset. 
MobileViTv3-1.0, 0.75 and 0.5 models outperform MobileViTv2-1.0, 0.75 and 0.5 models by 2.07\%, 1.73\% and 1.64\% respectively.

\begin{table}[!htb]
    \begin{subtable}[c]{0.5\textwidth}
    \scriptsize
    \centering
        \begin{tabular}{ccc}
        \toprule 
        {Backbone} & {\# Params. (M) $\downarrow$}& {mIOU(\%) $\uparrow$} \\ 
        \midrule
        MobileViTv1-XXS  & 1.90 & 73.60 \\
        \textbf{MobileViTv3-XXS}  & 1.96 & \textbf{74.04} (+0.44\%)   \\
        \midrule
        MobileViTv2-0.5  & 6.2 & 75.07   \\
        \textbf{MobileViTv3-0.5}  & 6.3 & \textbf{76.48} (+1.41\%)   \\
        \midrule
        MobileViTv1-XS  & 2.9 & 77.10 \\
        \textbf{MobileViTv3-XS}  & 3.3 & \textbf{78.77} (+1.60\%)   \\
        \midrule
        MobileViTv1-S  & 6.4 & 79.10 \\
        \textbf{MobileViTv3-S}  & 7.2 & \textbf{79.59} (+0.49\%)   \\
        \midrule
        MobileViTv2-1.0  & 13.32 & 78.94\\
        \textbf{MobileViTv3-1.0}  & 13.56 & \textbf{80.04} (+1.10\%)   \\
        \bottomrule
        \end{tabular}
        \subcaption{Segmentation on PASCAL VOC 2012 dataset}
        \label{table:segpascal}
    \end{subtable}
    \begin{subtable}[c]{0.5\textwidth}
    \scriptsize
    \centering
        \begin{tabular}{ccc}
        \toprule 
        {Backbone} & {\# Params. (M) $\downarrow$}& {mIOU(\%) $\uparrow$} \\ 
        \midrule
        MobileViTv2-0.5  & 6.31 & 31.93  \\
        \textbf{MobileViTv3-0.5}  & 6.37 & \textbf{33.57} (+1.64\%)  \\
        \midrule
        MobileViTv2-0.75  & 9.60 & 34.70 \\
        \textbf{MobileViTv3-0.75}  & 9.71 & \textbf{36.43} (+1.73\%)   \\
        
        \midrule
        MobileViTv2-1.0  & 13.40 & 37.06 \\
        \textbf{MobileViTv3-1.0} & 13.62 & \textbf{39.13} (+2.07\%)   \\
        \bottomrule
        \end{tabular}
        \subcaption{Segmentation on ADE20K dataset}
        \label{table:segade20k}
    \end{subtable}
\caption{Comparing MobileViTv3 segmentation task results on PASCAL VOC 2012 and ADE20K datasets. \# params of MobileViT models denotes the number of parameters in millions of the encoder/backbone architecture only.}
\end{table}

\subsection{Object detection} \label{results:obj}

\subsubsection{Implementation details}
MS-COCO dataset with 117K training and 5K validation images, is used to evaluate the detection performance of MobileViTv3 models. 
Similar to MobileViTv1, we integrated pretrained MobileViTv3 as a backbone network in Single Shot Detection network (SSD) \citep{liu2016ssd} and the standard convolutions in the SSD head are replaced with separable convolutions to create SSDLite network. 
SSDLite had also been used by other light-weight CNNs for evaluating performance on detection task. 
This SSDLite with pre-trained MobileViTv3-1.0, 0.75 and 0.5 is fine-tuned on MS-COCO dataset. 
Hyperparameters for training MobileViTv3-1.0, 0.75 and 0.5 are kept same as MobileViTv2-1.0, 0.75 and 0.5. 
Default hyperparameters include using images of input size 320 x 320, AdamW optimizer, weight decay of 0.05, cosine learning rate scheduler, total batch size of 128 (32 images per GPU), smooth L1 and cross-entropy losses are used for object localization and classification respectively. 
Training hyperparameters for MobileViTv3-S, XS and XXS are kept same as MobileViTv1-S, XS and XXS. 
Default hyperparameters include using images of input resolution of 320 x 320, AdamW optimizer, weight decay of 0.01, cosine learning rate scheduler, total batch size of 128 (32 images per GPU), smooth L1 and cross-entropy losses are used for object localization and classification respectively. 
Performance evaluation is done on validation set using mAP@IoU of 0.50:0.05:0.95 metric.

\subsubsection{Results}
Table \ref{table:detlight} and \ref{table:detheavy} show the detection results on COCO dataset. 
\# params of MobileViT models indicates number of parameters of the encoder/backbone architecture only. 
MobileViTv3 comparison with other light-weight CNN models is shown in Table \ref{table:detlight}.
MobileViTv3-XS outperforms MobileViTv1-XS by 0.8\% and MNASNet by 2.6\% mAP. 
Comparison with heavy-weight CNNs detailed in Table \ref{table:detheavy}. 
MobileViTv3-XS and MobileViTv3-1.0 surpasses MobileViTv1-XS and MobileViTv2-1.0 by 0.8\% and 0.5\% mAP respectively.

\begin{table}[!htb]
    \begin{subtable}[c]{0.5\textwidth}
        \centering
        \scriptsize
        \begin{tabular}{ccc} \toprule 
        Backbone & \# Params (M)$\downarrow$ & mAP(\%) $\uparrow$\\ \midrule
        MobileViTv1-XXS & 1.50 & 18.5\\
        \textbf{MobileViTv3-XXS} & 1.53 & \textbf{19.3} ($\uparrow$0.8\%)\\
        \midrule
        MobileViTv2-0.5 & 2 & 21.2  \\
        \textbf{MobileViTv3-0.5} & 2 & \textbf{21.8} ($\uparrow$0.6\%)  \\
        \midrule
        MobileViTv2-0.75 & 3.6 & 24.6  \\
        \textbf{MobileViTv3-0.75} & 3.7 & \textbf{25.0} ($\uparrow$0.4\%) \\
        \midrule
        MobileNetv3 & 4.9 & 22.0\\
        MobileNetv2 & 4.3 & 22.1\\
        MobileNetv1 & 5.1 & 22.2\\
        MixNet & 4.5 & 22.3\\
        MNASNet & 4.9 & 23.0 ($\uparrow$0.0\%)\\
        MobileViTv1-XS & 2.7 & 24.8 ($\uparrow$1.8\%)\\
        \textbf{MobileViTv3-XS} & 2.7 & \textbf{25.6} ($\uparrow$2.6\%)\\
        \bottomrule
        \end{tabular}
        \subcaption{Comparison w/light-weight CNNs}
        \label{table:detlight}
    \end{subtable}
    \begin{subtable}[c]{0.5\textwidth}
        \centering
        \scriptsize
        \begin{tabular}{ccc}
        \toprule 
        Backbone & \# Params (M)$\downarrow$ & mAP(\%) $\uparrow$\\
        \midrule 
        VGG & 35.6 & 25.1\\
        ResNet50 & 22.9 & 25.2 ($\uparrow$0.0\%)\\
        MobileViTv1-XS & 2.7 & 24.8 ($\downarrow$0.4\%)\\
        \textbf{MobileViTv3-XS} & 2.7 & \textbf{25.6} ($\uparrow$0.4\%)\\
        MobileViTv2-1.0 & 5.6 & 26.5 ($\uparrow$1.3\%) \\
        \textbf{MobileViTv3-1.0} & 5.8 & \textbf{27.0} ($\uparrow$1.8\%) \\
        MobileViTv1-S & 5.7 & 27.7 ($\uparrow$2.5\%)\\
        \textbf{MobileViTv3-S} & 5.5 & \textbf{27.3} ($\uparrow$2.1\%)\\
        
        \bottomrule
        \end{tabular}
        \subcaption{Comparison w/heavy-weight CNNs}
        \label{table:detheavy}
    \end{subtable}
    \caption{Comparing MobileViTv3 detection task results on COCO dataset with light-weight and heavy-weight CNNs. \# params of MobileViT models denotes the number of parameters of the encoder/backbone architecture only.}
\end{table}

\subsection{Improving Latency and Throughput} \label{results:latency}

\textbf{Implementation details}: We use GeForce RTX 2080 Ti GPU for obtaining latency timings.
Results are averaged over 10000 iterations. 
The timing results may vary $\pm$0.1 ms. 
Throughput for XXS, XS and S are calculated on 1000 iterations with batch size of 100. `Blocks' in Table \ref{table:latency} represents number of MobileViTv3-blocks in `layer4' of MobileViTv3 architectures (Table \ref{table1:fullarch}). \textit{To improve the latency, we reduce the number of MobileViT-blocks in `layer4' from 4 to 2}.

\textbf{Results}: Table \ref{table:latency} shows the latency and throughput results.
MobileViTv3-XXS with similar parameters and FLOPs as the baseline MobileViTv1-XXS, along with 1.98\% accuracy improvement achieves similar latency of $\sim$7.1 ms. 
MobileViTv3-XXS with two MobileViT-blocks instead of four, has 30\% less FLOPs and achieves latency of 6.24 ms which is $\sim$1 ms faster than the baseline MobileViTv1-XXS. 
With similar changes in MobileViTv3-XS and MobileViTv3-S architecture, FLOPs are reduced by 13.5\% and 17.82\% respectively and latency is reduced by $\sim$1 ms and $\sim$0.7 ms respectively.

\begin{table}[!htb]
\centering
\scriptsize
\begin{tabular}{ccccccc} \toprule
    {Model} & {Blocks ($\downarrow$)} & {FLOPs (M)$\downarrow$} & {\# Params (M)$\downarrow$}  & {Top-1 (\%)$\uparrow$} &  {Throughput ($\uparrow$)}  & {\# Time (ms) ($\downarrow$)}  \\ 
    \midrule
    MobileViTv1-XXS & 4  & 364 & 1.30 & 69.00 & 2124 & 7.24  \\
    MobileViTv3-XXS & 4  & 289 & 1.25 & 70.98 ($\uparrow$1.98\%) & 2146 & 7.12  \\
    \textbf{MobileViTv3-XXS} & \textbf{2}  & 256 ($\downarrow$30\%) & 1.14 & \textbf{70.23} ($\uparrow$1.23\%) & \textbf{2308} & \textbf{6.24}  \\
    \midrule
    MobileViTv1-XS & 4  & 986 & 2.3 & 74.80 & 1097 & 7.32  \\
    MobileViTv3-XS & 4  & 927 & 2.5 & 76.7 ($\uparrow$1.9\%)  & 1078 & 7.20  \\
    \textbf{MobileViTv3-XS} & \textbf{2}  & 853 ($\downarrow$13.5\%) & 2.3 & \textbf{76.1}  ($\uparrow$1.3\%) & \textbf{1129} & \textbf{6.35}  \\
    \midrule
    MobileViTv1-S & 4  & 2009 & 5.6 & 78.40 & 822 & 7.34  \\
    MobileViTv3-S & 4  & 1841 & 5.8 & 79.3 ($\uparrow$0.9\%)  & 824 & 7.29  \\
    \textbf{MobileViTv3-S} & \textbf{2}  & 1651 ($\downarrow$17.82\%) & 5.2 & \textbf{79.06}  ($\uparrow$0.6\%) & \textbf{876} & \textbf{6.60}  \\
    \bottomrule
\end{tabular}
\caption{Latency (in milliseconds) and throughput (in Images per sec) comparison between MobileViTv3-XXS, XS, and S and MobileViTv1-XXS, XS, and S. While keeping the parameters and Top-1 accuracy similar to MobileViTv1, MobileViTv3 with 2 blocks reduces the number of FLOPs and improves the throughput and latency.}
\label{table:latency}
\end{table}

\subsection{Ablation study of our proposed MobileViTv3 block}

\subsubsection{Implementation details}
We study the effect of the four proposed changes on MobileViTv1-S block by adding changes one by one.
The final model with all the four changes is our unscaled version and we name it: MobileViTv3-S(unscaled). 
To match the number of parameters of MobileViTv1 we increase the width of MobileViTv3-S(unscaled), giving us MobileViTv3-S.
Top-1 accuracy on ImageNet-1K of each change is recorded and compared to other proposed changes. 
In this ablation study we train models for 100 epochs, use batch size of 192 (32 images per GPU) and other hyper-parameters are default as given in section \ref{section:classimpdetail}. All the proposed changes in our work are applied in the MobileViTv1-block which is made up of local representation, global representation and fusion blocks. 
In Table \ref{table:ablation}, `conv-3x3' represents of 3x3 convolutional layer in fusion block, `conv-1x1' represents of 1x1 convolutional layer in fusion block, `Input-Concat' represents concatenating input features with global representation in the fusion block, `Local-Concat' represents concatenating local-representation block output features with global representation in the fusion block, `Input-Add' represents adding input features to the output of fusion block, `DWConv' represents using depthwise convolutional layer in the local representation block instead of a normal convolutional layer and `Top-1' represents Top-1 accuracy on ImageNet-1K dataset.

\subsubsection{With 100 training epochs}
Results are shown in Table \ref{table:ablation}. 
The baseline MobileViTv1-S, in fusion block, concatenates input features with global representation block features and uses 3x3 convolutional layer. 
Also, it uses normal 3x3 convolutional layer in the local representation block. 
This baseline achieves an accuracy of 73.7\%. 

\textbf{Replacing 3x3 convolution with 1x1 convolutional layer in fusion block}, MobileViTv3-S(unscaled) achieves 1.1\% improvement. 
This result supports the assumption that simplifying the fusion block's task (allowing fusion layer to fuse local and global features independent of the other location's local and global features) should help optimization to attain better performance. 
Along with 1x1 convolutional layer in fusion block, \textbf{concatenating local representation features instead of input features} results in similar performance gains of 1\% compared to concatenating input features. 
This allows us to incorporate the next change i.e, \textbf{to add input features to the output of fusion block} to create a residual connection for helping optimization of deeper layers in the model.
With this change, MobileViTv3-S(unscaled) attains 1.6\% accuracy gain over the baseline MobileViTv1-S and 0.6\% gain over the last change demonstrating the clear advantage of this residual connection. 
To further reduce number of parameters and FLOPs in MobileViTv3-block, \textbf{depth-wise convolutional layer is used instead of normal convolutional layer in the local representation block}. MobileViTv3-S(unscaled) maintains high accuracy gains by achieving 1.3\% gain over the baseline. 0.3\% accuracy drop can be observed when compared to the previous change. 
We adopt this change since it reduces parameters and FLOPs without significantly impacting performance and helps in scaling of MobileViTv3-block.

\begin{table}[!htb]
\centering
\scriptsize
\begin{tabular}{cccccccc} \toprule
    {Model} & {Conv 3x3} & {Conv 1x1} & {Input Concat}& {Local Concat} & {Input Add} & {DW Conv} & {Top-1 (\%)$\uparrow$}  \\ 
    \arrayrulecolor{black}\midrule
    MobileViTv1-S  & \checkmark  &  & \checkmark  &    &    &    & 73.7 ($\uparrow$0.0\%) \\
    \arrayrulecolor{lightgray}\midrule
    MobileViTv3-S (unscaled)  &    & \checkmark  & \checkmark  &    &    &    & 74.8 ($\uparrow$1.1\%) \\
    \midrule
    MobileViTv3-S (unscaled)  &    & \checkmark  &    & \checkmark  &    &    & 74.7 ($\uparrow$1.0\%)  \\
    \midrule
    MobileViTv3-S (unscaled)  &    & \checkmark  &    & \checkmark  & \checkmark  &    & 75.3 ($\uparrow$1.6\%) \\
    \midrule
    MobileViTv3-S (unscaled)  &    & \checkmark  &    & \checkmark  & \checkmark  & \checkmark  & 75.0 ($\uparrow$1.3\%)\\
    \arrayrulecolor{black}\bottomrule
\end{tabular}
\caption{ Ablation study of MobileViTv3 block. `unscaled' indicates that the number of channels in the architecture are kept same as the baseline MobileViTv1. \checkmark represents incorporating the change in the block.}
\label{table:ablation}
\end{table}

\subsubsection{With 300 training epochs}
When trained for 300 epochs with the batch size of 192, the baseline MobileViTv1-S achieves Top-1 accuracy of 75.6\%, which is lower by 2.8\% compared to reported accuracy on MobileViTv1-S trained on 1024 batch size. Results shown in Table \ref{table:300ablation}.
With all the four proposed changes implemented in MobileViTv1-S architecture to form MobileViTv3-S(unscaled), the model reaches Top-1 accuracy of 77.5\%, which outperforms the baseline by 1.9\% with 22.7\% and 18.6\% less parameters FLOPs respectively. 
 
\begin{table}[!htb]
\centering
\scriptsize
\begin{tabular}{cccccc} \toprule
    {Model} & {Training batch size} & {FLOPs (M)$\downarrow$} & {\# Params (M)$\downarrow$} & {Top-1 (\%)$\uparrow$}  \\ 
    \midrule
    MobileViTv1-S  & 192  & 2009 & 5.6 & 75.6  \\
    MobileViTv3-S(unscaled)  & 192  & 1636 ($\downarrow$18.6\%)  & 4.3 ($\downarrow$22.7\%) & 77.5 ($\uparrow$1.9\%)   \\
    \midrule
    MobileViTv1-S  & 1024  & 2009 & 5.6 & 78.4  \\
    MobileViTv3-S  & 384  & 1841 ($\downarrow$8.3\%)  & 5.8 ($\uparrow$3.6\%) & 79.3 ($\uparrow$0.9\%)   \\
    \bottomrule
\end{tabular}
\caption{MobileViTv3-S(unscaled), MobileViTv1-S and MobileViTv3-S Top-1 ImageNet-1K accuracy comparisons. With similar parameters and FLOPs after scaling, MobileViTv3-S is able to exhibit better performance than baseline MobileViTv1-S.}
\label{table:300ablation}
\end{table}

 MobileViTv3-S(unscaled) architecture though better than the baseline MobileViTv1-S with training batch size of 192, performs worse than the MobileViTv1-S trained at batch size of 1024. 
 Therefore, MobileViTv3-S, XS and XXS models are scaled to have similar parameters and FLOPs as MobileViTv1-S, XS and XXS and are trained with batch size of 384. 
 Table \ref{table:300ablation} demonstrates that after scaling, MobileViTv3-S is able to outperform MobileViTv1-S by achieving 79.3\% accuracy with similar parameters and FLOPs.
 Table \ref{table:comp_batch} shows that MobileViTv3-XS and XXS are also able to surpass MobileViTv3-XS and XXS performance by 1.9\% and 2.0\% respectively with similar parameters and FLOPs. 

\section{Discussion and limitations}
This work is an effort towards improving performance of models for resource constrained environments like mobile phones.
We looked at reducing memory (parameters), computation (FLOPs), latency while boosting accuracy and throughput.
With the proposed changes to MobileViT blocks we achieve higher accuracy, with same memory and computation as the baseline MobileViTv1 and v2 as seen in section \ref{results:imgnet}. Table \ref{table:comp_batch} shows fine-tuning results which also outperform the fine-tuned MobileViTv2 models.
Section \ref{results:latency} shows how we can achieve better latency and throughput with minimal impact on the accuracy of the model.
While MobileViTv3 has higher accuracy and lower or similar parameters as compared to other mobile-CNNs, it's higher FLOPs can be an issue for edge devices (Figure \ref{fig:comp_cnn}).
This limitation of MobileViTv3 architecture is inherited from the self-attention module of ViTs.
To solve this issue, we will further explore optimization of the self-attention block.
Table \ref{table:comp_batch} shows results on Imagenet-1K. The reported accuracies of MobileViTv3-XXS, XS and S models on Imagenet-1K can potentially be further improved by increasing the training batch size to 1024 similar to the baseline model.
The proposed fusion of input features, local features (CNN features) and global features (ViT features) shown in this paper can also be explored in other hybrid architectures.

\subsubsection*{Acknowledgments}
We thank Micron Deep Learning Accelerator (MDLA) team for support with the training infrastructure, feedback and discussions.

\bibliography{mobilevit_v3}
\bibliographystyle{iclr2022_conference}

\appendix

\section{Object detection and Semantic segmentation results}

\subsection{Object detection on COCO dataset}

\begin{figure}[ht]
\centering
\begin{subfigure}{.5\textwidth}
  \centering
  \includegraphics[width=0.9\linewidth]{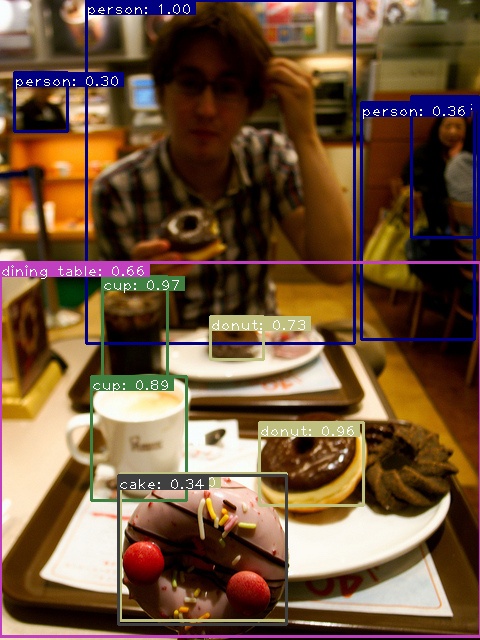}
\end{subfigure}\hfil
\begin{subfigure}{.5\textwidth}
  \centering
  \includegraphics[width=0.9\linewidth]{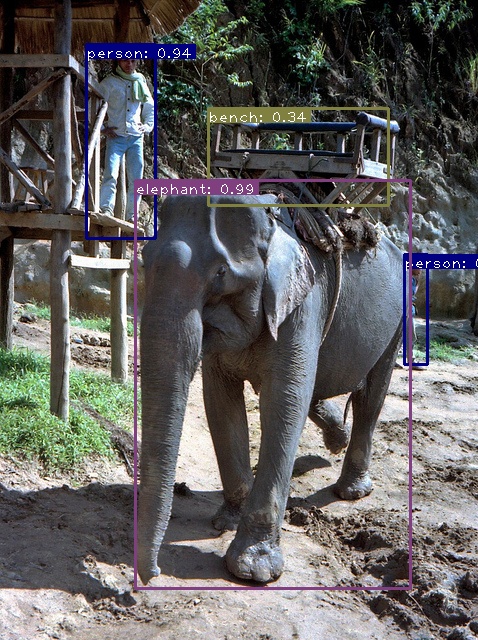}
\end{subfigure}\hfil

\vspace{0.5cm}

\begin{subfigure}{.5\textwidth}
  \centering
  \includegraphics[width=0.9\linewidth]{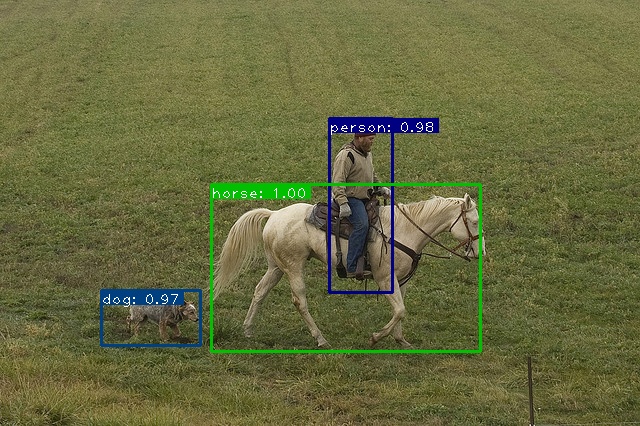}
\end{subfigure}\hfil
\begin{subfigure}{.5\textwidth}
  \centering
  \includegraphics[width=0.9\linewidth,trim={0 1.5cm 0 0},clip]{allfigures/imgs-det-v1/000000050331.jpg}
\end{subfigure}\hfil

\vspace{0.5cm}

\begin{subfigure}{.5\textwidth}
  \centering
  \includegraphics[width=0.9\linewidth]{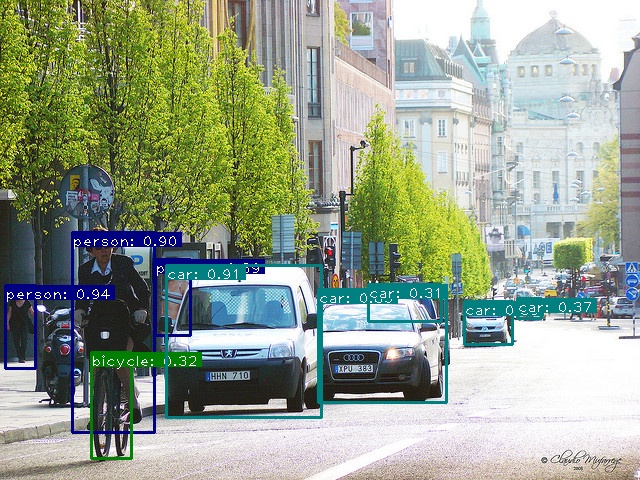}
\end{subfigure}\hfil
\begin{subfigure}{.5\textwidth}
  \centering
  \includegraphics[width=0.9\linewidth]{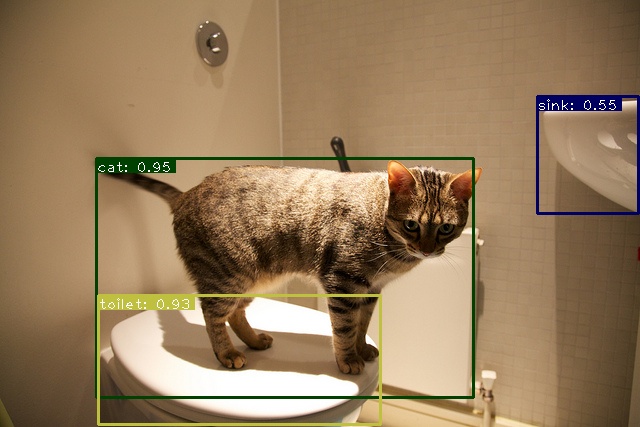}
\end{subfigure}\hfil
\caption{Object detection results using SSD-Lite model with MobileViTv3-S as its backbone.}
\label{fig:det-mv3v1}
\end{figure}

\begin{figure}[ht]
\centering
\begin{subfigure}{.5\textwidth}
  \centering
  \includegraphics[width=0.9\linewidth]{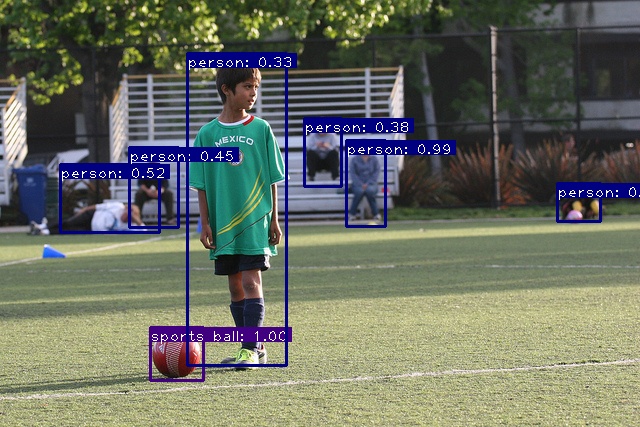}
\end{subfigure}\hfil
\begin{subfigure}{.5\textwidth}
  \centering
  \includegraphics[width=0.9\linewidth]{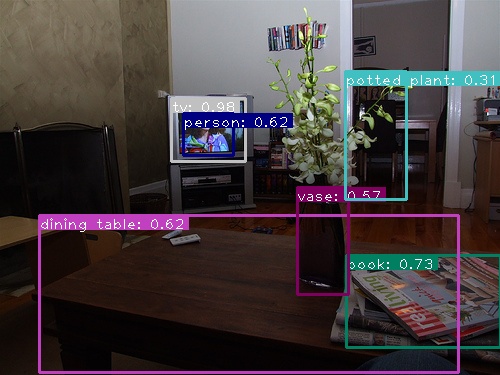}
\end{subfigure}\hfil

\vspace{0.5cm}

\begin{subfigure}{.5\textwidth}
  \centering
  \includegraphics[width=0.9\linewidth]{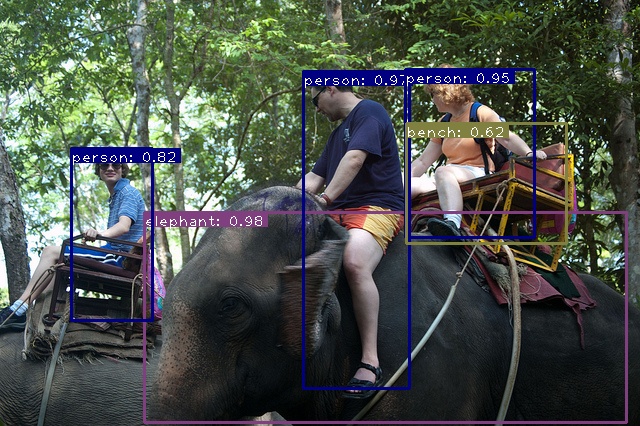}
\end{subfigure}\hfil
\begin{subfigure}{.5\textwidth}
  \centering
  \includegraphics[width=0.9\linewidth,trim={0 0 0 0},clip]{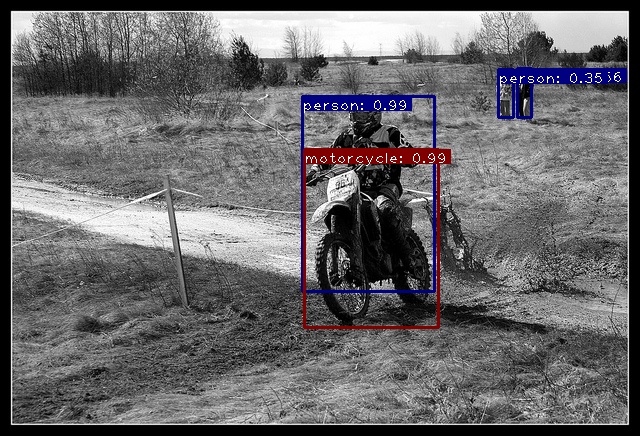}
\end{subfigure}\hfil

\vspace{0.5cm}

\begin{subfigure}{.5\textwidth}
  \centering
  \includegraphics[width=0.9\linewidth]{allfigures/imgs-det-v2/000000000724.jpg}
\end{subfigure}\hfil
\begin{subfigure}{.5\textwidth}
  \centering
  \includegraphics[width=0.9\linewidth]{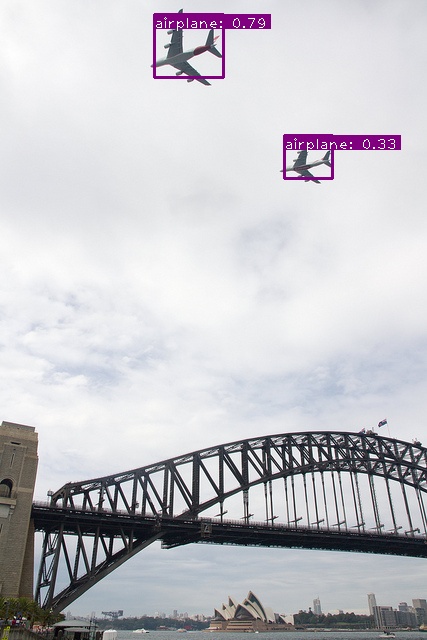}
\end{subfigure}\hfil
\caption{Object detection results using SSD-Lite model with MobileViTv3-1.0 as its backbone.}
\label{fig:det-mv3v2}
\end{figure}

Figure \ref{fig:det-mv3v1} shows object detection results on COCO validation images using SSD-Lite with MobileViTv3-S as its backbone.
Figure \ref{fig:det-mv3v2} shows object detection results on COCO validation images using SSD-lite with MobileViTv3-1.0 as its backbone.
The images shown in figure \ref{fig:det-mv3v1} include challenging object detection examples (blurred human/person and complex background).

\subsection{Semantic segmentation on PascalVOC2012 dataset}

\begin{figure}[ht]
\centering
\begin{subfigure}{.33\textwidth}
  \centering
  \includegraphics[width=0.9\linewidth]{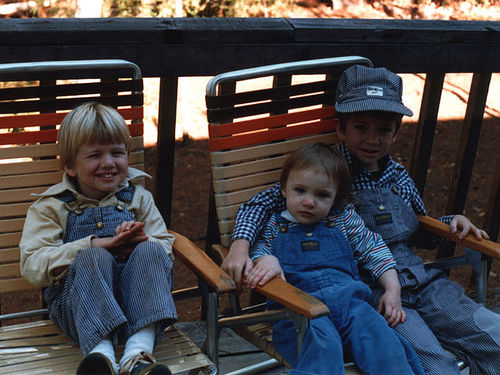}
\end{subfigure}\hfil
\begin{subfigure}{.33\textwidth}
  \centering
  \includegraphics[width=0.9\linewidth]{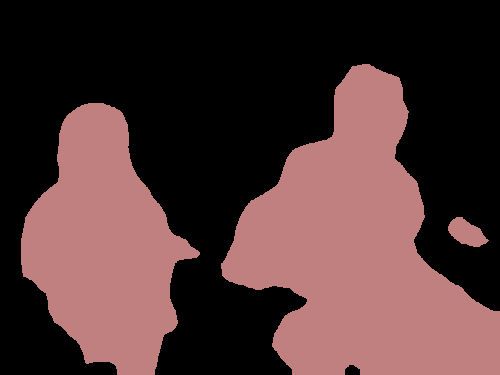}
\end{subfigure}\hfil
\begin{subfigure}{.33\textwidth}
  \centering
  \includegraphics[width=0.9\linewidth]{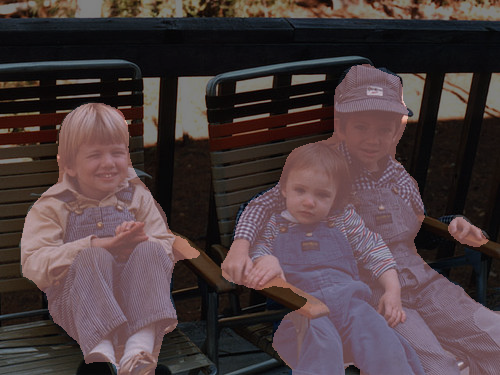}
\end{subfigure}\hfil

\vspace{0.5cm}

\begin{subfigure}{.33\textwidth}
  \centering
  \includegraphics[width=0.9\linewidth]{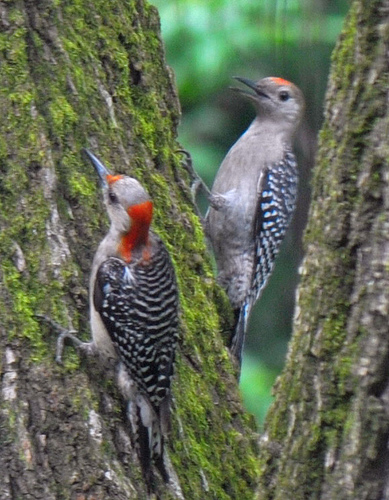}
\end{subfigure}\hfil
\begin{subfigure}{.33\textwidth}
  \centering
  \includegraphics[width=0.9\linewidth]{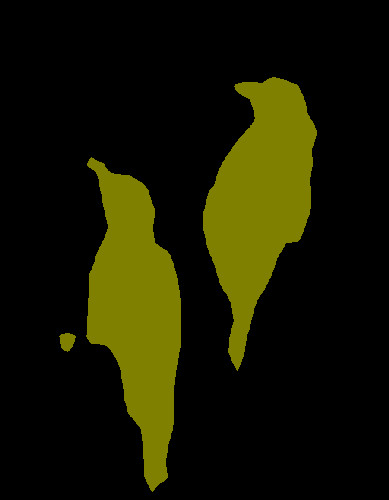}
\end{subfigure}\hfil
\begin{subfigure}{.33\textwidth}
  \centering
  \includegraphics[width=0.9\linewidth]{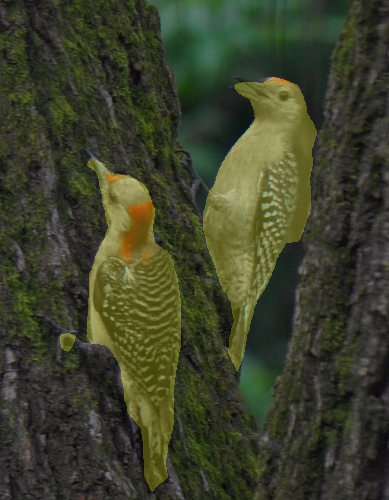}
\end{subfigure}\hfil

\vspace{0.5cm}

\begin{subfigure}{.33\textwidth}
  \centering
  \includegraphics[width=0.9\linewidth]{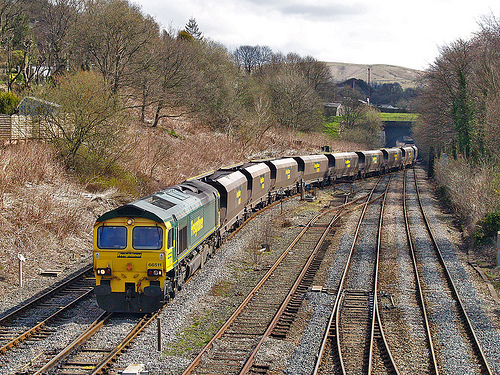}
\end{subfigure}\hfil
\begin{subfigure}{.33\textwidth}
  \centering
  \includegraphics[width=0.9\linewidth]{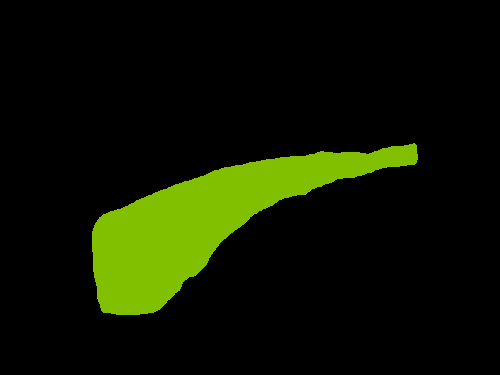}
\end{subfigure}\hfil
\begin{subfigure}{.33\textwidth}
  \centering
  \includegraphics[width=0.9\linewidth]{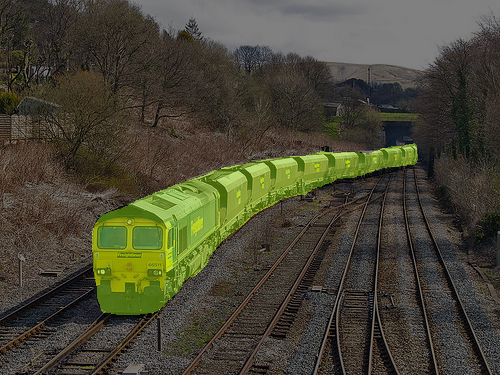}
\end{subfigure}\hfil

\vspace{0.5cm}

\begin{subfigure}{.33\textwidth}
  \centering
  \includegraphics[width=0.9\linewidth]{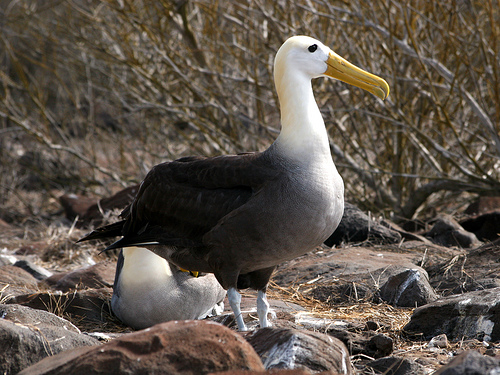}
\end{subfigure}\hfil
\begin{subfigure}{.33\textwidth}
  \centering
  \includegraphics[width=0.9\linewidth]{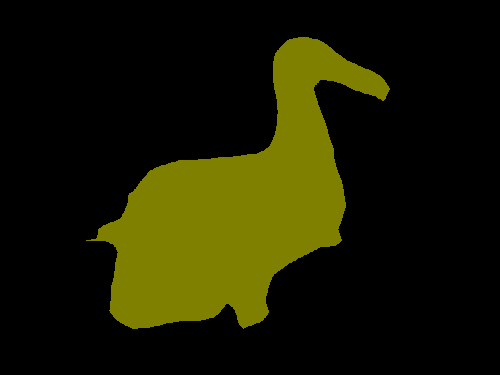}
\end{subfigure}\hfil
\begin{subfigure}{.33\textwidth}
  \centering
  \includegraphics[width=0.9\linewidth]{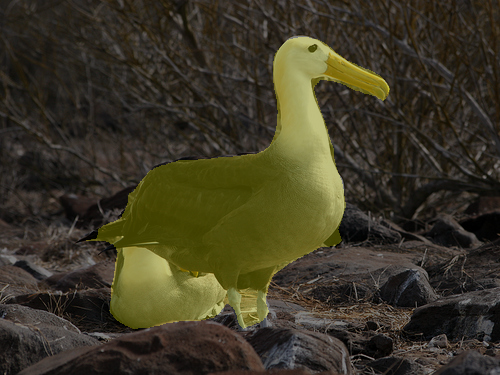}
\end{subfigure}\hfil
\caption{Semantic segmentation results using Deeplabv3 with MobileViTv3-S as its backbone.}
\label{fig:seg-mv3v1}
\end{figure}

\begin{figure}[ht]
\centering
\begin{subfigure}{.33\textwidth}
  \centering
  \includegraphics[width=0.9\linewidth]{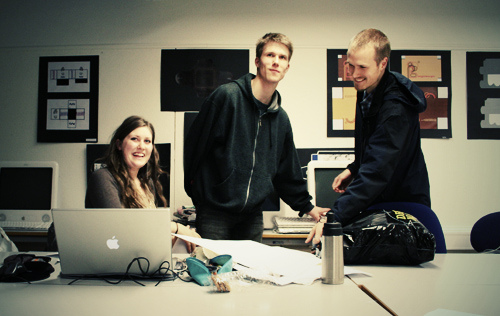}
\end{subfigure}\hfil
\begin{subfigure}{.33\textwidth}
  \centering
  \includegraphics[width=0.9\linewidth]{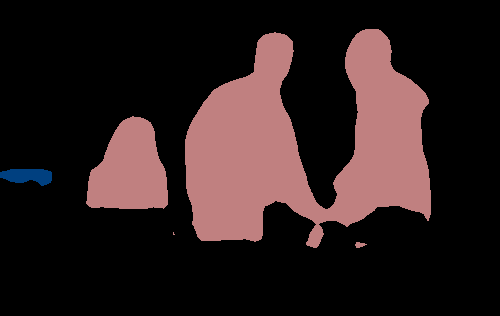}
\end{subfigure}\hfil
\begin{subfigure}{.33\textwidth}
  \centering
  \includegraphics[width=0.9\linewidth]{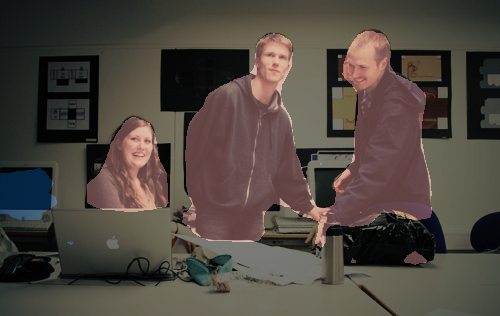}
\end{subfigure}\hfil

\vspace{0.5cm}

\begin{subfigure}{.33\textwidth}
  \centering
  \includegraphics[width=0.9\linewidth]{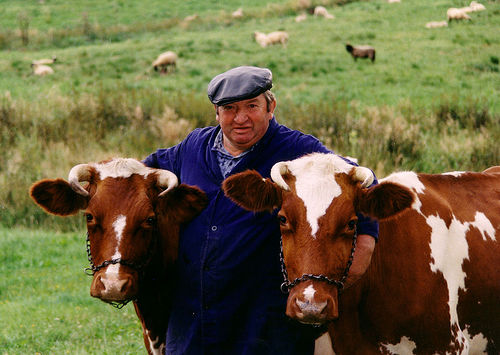}
\end{subfigure}\hfil
\begin{subfigure}{.33\textwidth}
  \centering
  \includegraphics[width=0.9\linewidth]{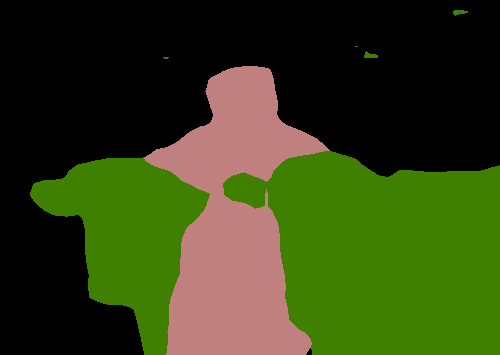}
\end{subfigure}\hfil
\begin{subfigure}{.33\textwidth}
  \centering
  \includegraphics[width=0.9\linewidth]{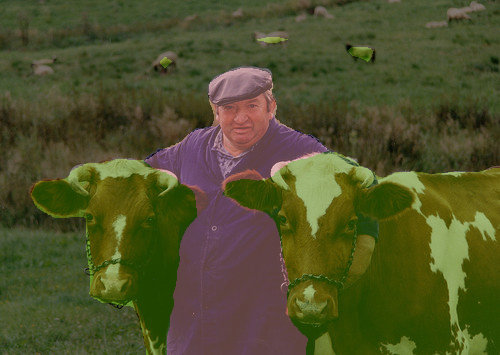}
\end{subfigure}\hfil

\vspace{0.5cm}

\begin{subfigure}{.33\textwidth}
  \centering
  \includegraphics[width=0.9\linewidth]{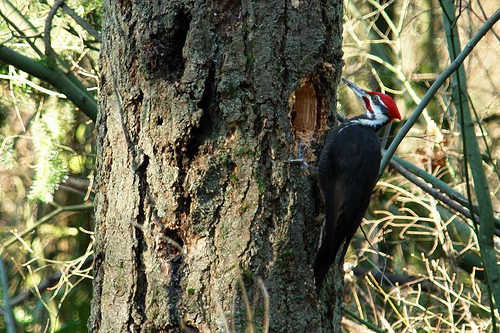}
\end{subfigure}\hfil
\begin{subfigure}{.33\textwidth}
  \centering
  \includegraphics[width=0.9\linewidth]{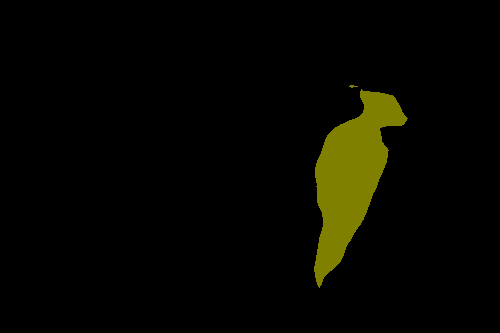}
\end{subfigure}\hfil
\begin{subfigure}{.33\textwidth}
  \centering
  \includegraphics[width=0.9\linewidth]{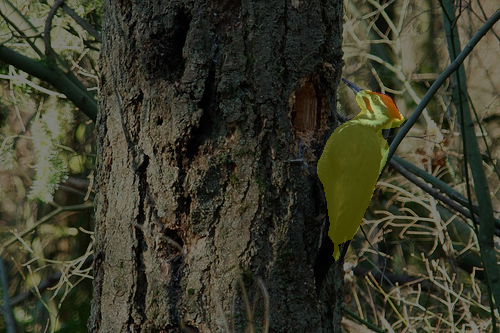}
\end{subfigure}\hfil

\vspace{0.5cm}

\begin{subfigure}{.33\textwidth}
  \centering
  \includegraphics[width=0.9\linewidth]{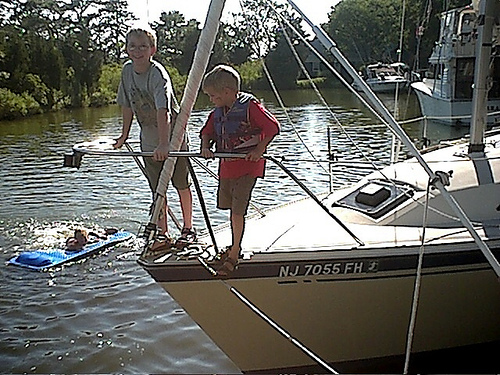}
\end{subfigure}\hfil
\begin{subfigure}{.33\textwidth}
  \centering
  \includegraphics[width=0.9\linewidth]{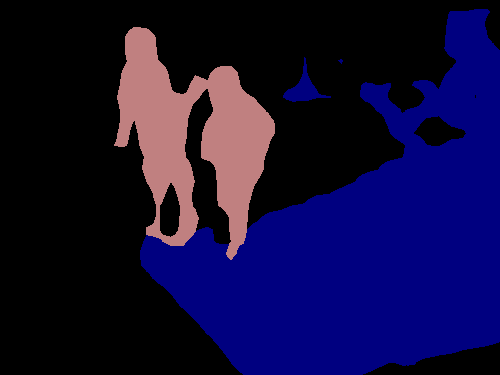}
\end{subfigure}\hfil
\begin{subfigure}{.33\textwidth}
  \centering
  \includegraphics[width=0.9\linewidth]{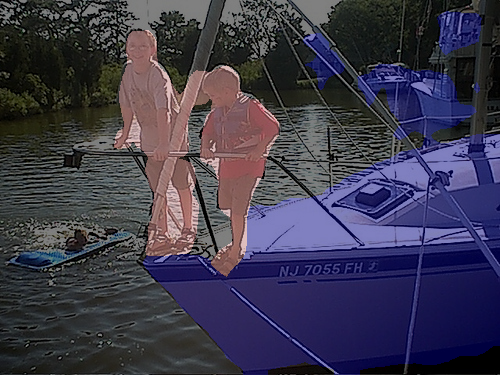}
\end{subfigure}
\caption{Semantic segmentation results using Deeplabv3 with MobileViTv3-1.0 as its backbone.}
\label{fig:seg-mv3v2}
\end{figure}

Figure \ref{fig:seg-mv3v1} shows segmentation results on PascalVOC2012 validation images using Deeplabv3 with MobileViTv3-S as its backbone.
In figure \ref{fig:seg-mv3v1} and \ref{fig:seg-mv3v2}, moving from left to right we provide the input image, the corresponding segmentation output, and the overlay of segmentation output on the input image.

\end{document}